\documentclass{article}
\usepackage{spconf}
\usepackage{amsmath,graphicx}
\usepackage{times}
\usepackage{epsfig}
\usepackage{graphicx}
\usepackage{amsmath}
\usepackage{amssymb}
\usepackage{algorithm}
\usepackage{algorithmic}
\usepackage{enumerate}
\usepackage{multirow}
\usepackage{multicol}
\usepackage{arydshln}
\usepackage{color}
\usepackage{amssymb}

\def\x{{\mathbf x}}
\def\0{{\mathbf 0}}
\def\1{{\mathbf 1}}

\def\l{{\mathbf l}}

\def\v{{\mathbf v}}

\def\x{{\mathbf x}}
\def\y{{\mathbf y}}

\def\D{{\mathbf D}}
\def\E{{\mathbf E}}

\def\L{{\mathbf L}}

\def\V{{\mathbf V}}
\def\W{{\mathbf W}}

\def\ie{{\textit{i.e.}}}

\def\cE{{\mathcal E}}

\def\cG{{\mathcal G}}

\def\cN{{\mathcal N}}
\def\cO{{\mathcal O}}

\def\balpha{{\boldsymbol \alpha}}

\title{Graph Sparsification for GCN Towards Optimal Crop Yield Predictions}
\name{Saghar Bagheri$^{\dag}$, Gene Cheung$^{\dag}$\thanks{The work of G. Cheung was supported in part by the Natural Sciences and Engineering Research
Council of Canada (NSERC) RGPIN-2019-06271, RGPAS-2019-00110.}, Tim Eadie$^\star$}
\address{$^\dag$York University, Canada~~~~~
$^\star$Growers Edge}
\begin{document}
%\ninept
%
\maketitle
\begin{abstract}
In agronomics, predicting crop yield at a per field / county granularity is important for farmers to minimize uncertainty and plan seeding for the next crop cycle. 
While state-of-the-art prediction techniques employ graph convolutional nets (GCN) to predict future crop yields given relevant features and crop yields of previous years, a dense underlying graph kernel requires long training and execution time. 
In this paper, we propose a graph sparsification method based on the Fiedler number to remove edges from a complete graph kernel, in order to lower the complexity of GCN training / execution. 
Specifically, we first show that greedily removing an edge at a time that induces the minimal change in the second eigenvalue leads to a sparse graph with good GCN performance.
We then propose a fast method to choose an edge for removal per iteration based on an eigenvalue perturbation theorem.
Experiments show that our Fiedler-based method produces a sparse graph with good GCN performance compared to other graph sparsification schemes in crop yield prediction. 
\end{abstract}
\vspace{-0.05in}
\section{Introduction}
\label{sec:intro}
\vspace{-0.05in}
As a fundamental problem in agronomics, crop yield prediction at a per field / county granularity has been intensively studied in the past two decades \cite{Cai17}. 
Early attempts employed model-based \textit{machine learning} (ML) methods such as LASSO regression, linear regression, and random forest for prediction. 
%One notable and widely used model is XGBoost \cite{TChen16}, a generic tree boosting system used for many machine learning tasks. 
With the recent advent of deep learning, powerful deep network architectures like \textit{convolutional neural nets} (CNN) and \textit{recurrent neural nets} (RNN) have been used \cite{Sun19,Shahhosseini21}. 
However, given that the underlying data kernel (a geographic network of crop-growing fields or counties) is irregular, CNN that typically operates on regular kernels like an image grid does not perform well.
Instead, \textit{graph convolutional nets} (GCN) that perform filtering on graph kernels describing pairwise similarities are more suitable \cite{kipf17}. 
We focus on the study of graph kernels for GCN in this work.

While a \textit{complete} graph $\cG$---there exists an edge $(i,j)$ from any node $i$ to any other node $j$ with weight $w_{i,j}$ encoding their pairwise similarity---is clearly the most informative graph kernel\footnote{Note that a complete graph can lead to \textit{oversmoothing} faster than a sparse graph if the GCN contains too many layers, resulting in worse performance. This is an orthogonal problem and is studied in works such as \cite{oono2020graph,zeng23}.}, representing $\cG$ as a \textit{dense} Laplacian matrix $\L$ means long GCN training and execution time.
Conventionally, in the \textit{graph signal processing} (GSP) literature \cite{ortega18ieee,cheung18} a \textit{sparse} graph is chosen instead via simple sparsification methods like $k$ Nearest Neighbors (kNN) and $\epsilon$-thresholding (to be described in details in Section\;\ref{subsec:sparseMethod}).
However, kNN often removes edges with large weights that are important for effective filtering, while $\epsilon$-thresholding often disconnects the graph into sub-graphs or isolated nodes.
Though there exist graph learning works \cite{hu20,yang22} that optimize the computation of edge weights $\{w_{i,j}\}$ given pre-computed per-node features, the sparsification of complete graphs towards complexity reduction of GCN training is relatively unexplored.

In this paper, we propose a new graph sparsification method based on the \textit{Fiedler number}---the second smallest eigenvalue $\lambda_2$ of the Laplacian matrix $\L$. 
The Fiedler number is known to quantify the ``connectedness" of the underlying graph \cite{Chung1997}. 
For example, if $\cG$ is disconnected, then $\lambda_2 = 0$, and if $\cG$ is an unweighted complete graph, then $\lambda_2 = N$, the number of graph nodes.
Thus, to maximally preserve graph connectedness, we seek per iteration to remove an edge whose removal would induce the minimal change in $\lambda_2$. 
We first empirically demonstrate this greedy edge removal method by Fiedler number does lead to good GCN performance.
We then propose a fast method to choose edges for removal based on an eigenvalue perturbation theorem \cite{ipsen09}.
Experiments show that our Fiedler-based method produces a sparse graph kernel with good GCN performance, while reducing the GCN training time significantly.

%\section{Related Work}
%\label{sec:related work}
%\input{Related Work.tex}

\vspace{-0.05in}
\section{Preliminaries}
\label{sec:prelim}
\vspace{-0.05in}
\subsection{GSP Definitions}

\vspace{-0.05in}
A graph $\cG(\cN,\cE,\W)$ contains a node set $\cN = \{1, \ldots, N\}$ of $N$ nodes and an edge set $\cE$, where edge $(i,j) \in \cE$ has weight $w_{i,j} = W_{i,j}$ specified by the $(i,j)$-th entry of the \textit{adjacency matrix} $\W$. 
If $(i,j) \not\in \cE$, then $W_{i,j} = 0$. 
We assume $\cG$ has no self-loops, and thus $W_{i,i} = 0, \forall i$.
We assume an undirected graph, and thus $W_{i,j} = W_{j,i}$ and $\W$ is symmetric.
A diagonal \textit{degree matrix} $\D \in \mathbb{R}^{N \times N}$ has diagonal entries $D_{i,i} = \sum_{j} W_{i,j}$.
Finally, a \textit{combinatorial graph Laplacian matrix} is defined as $\L \triangleq \D - \W$ \cite{ortega18ieee}. 
For a \textit{positive} graph where $w_{i,j} \geq 0, \forall i,j$, it can be proven \cite{cheung18} that $\L$ is \textit{positive semi-definite} (PSD), \ie, $\x^\top \L \x \geq 0, \forall \x \in \mathbb{R}^N$, or equivalent, all eigenvalues $\{\lambda_k\}$ of $\L$ are non-negative. 

\vspace{-0.05in}
\subsection{Graph Filters}

\vspace{-0.05in}
The $k$-th eigen-pair $(\lambda_k, \v_k)$ of Laplacian $\L$, \ie, $\L \v_k = \lambda_k \v_k$, is conventionally interpreted as the $k$-th graph frequency and Fourier mode, respectively \cite{ortega18ieee}. 
A signal $\x \in \mathbb{R}^N$ can be spectrally decomposed into its frequency components via $\balpha = \V^\top \x$, where matrix $\V \in \mathbb{R}^{N \times N}$ contains eigenvectors $\{\v_k\}$ as columns, and $\alpha_k = \v_k^\top \x$ is the coefficient for the $k$-th frequency. 
A $P$-tap \textit{graph filter} $P(\L) = \sum_{p=0}^P a_p \L^p$ is a polynomial of Laplacian $\L$ with filter coefficients $\{a_p\}_{p=0}^P$, so that $P(\L) \x$ filters signal $\x$ via graph frequency selection / attenuation. 
Each convolutional filter in a neural layer of a GCN can be viewed as a graph filter.
Thus, a denser $\L$ would mean more computation for the GCN.

\vspace{-0.05in}
\section{Graph Sparsification}
\label{sec:Method}
\vspace{-0.05in}
\subsection{Methods for Graph Sparsification}
\label{subsec:sparseMethod}

\vspace{-0.05in}
We overview candidate methods using which a complete graph can be sparsified.

\vspace{-0.05in}
\subsubsection{$k$ Nearest Neighbors}

\vspace{-0.05in}
One common method to sparsify a complete graph $\cG$ with all possible edges is \textit{$k$ nearest neighbors} (kNN), 
Specifically, each node $i$ keeps only the $k$ connected edges $(i,j) \in \cE$ with the largest edge weights $w_{i,j}$ and removes the rest (if possible). 
This procedure must be done symmetrically, so that if $w_{i,j}$ is one of $k$ largest edge weights from node $j$, then edge $(i,j)$ is not removed.
This results in a graph where each node has at least $k$ connected edges. 
The advantage of kNN is that the sparsified graph remains connected.
The disadvantage is that some edges with large weights $w_{i,j}$ that are pertinent to effective filtering may be removed.

\vspace{-0.05in}
\subsubsection{$\epsilon$-Thresholding}

\vspace{-0.05in}
Another common method to sparsify a complete graph $\cG$ is \textit{$\epsilon$-thresholding}: remove all edges $(i,j) \in \cE$ with weights $w_{i,j} < \epsilon$, where $\epsilon \in \mathbb{R}_+$ is a chosen positive parameter. 
The advantage of $\epsilon$-thresholding is that edges with large weights are preserved.
The disadvantage is that some nodes may be left with very few edges; in some extreme cases, nodes can be disconnected from the rest of the graph, preventing filtering between disconnected nodes.

\vspace{-0.05in}
\subsubsection{Hybrid kNN / $\epsilon$-Thresholding}

\vspace{-0.05in}
We can also sparsify a complete graph $\cG$ by combining kNN and $\epsilon$-thresholding: we first remove edges from each node $i$ until it has an upper bound of $d_{\max}$ edges, then edges are further removed if i) $w_{i,j} < \epsilon$; and ii) the number of edges $(i,j) \in \cE$ connected to $i$ remains more than $d_{\min}$. 
By appropriately selecting $d_{\max}$, $d_{\min}$ and $\epsilon$, each node maintains a minimum of $d_{\min}$ connections to the graph, while edges $(i,j)$ with small weights $w_{i,j} < \epsilon$ are removed. 

\vspace{-0.05in}
\subsubsection{All-Pair Shortest Paths}

\vspace{-0.05in}
One can remove one edge at a time from a complete graph $\cG$ based on a metric evaluating the ``connectedness" of the graph.
One candidate metric is \textit{all-pair shortest paths} (APSP): we compute the sum of shortest paths from any node $i \in \cN$ to another node $j \in \cN$ such that $j \neq i$. 
APSP can be computed using the known Floyd-Warshall algorithm in $\cO(N^3)$ complexity \cite{floyd62}.
In each iteration, we select the edge that induces the smallest change in APSP for removal.

\vspace{-0.05in}
\subsubsection{Fiedler Number}
\label{subsubsec:Fiedler}

\vspace{-0.05in}
An alternative metric to quantify the connectedness of a graph is the second smallest eigenvalue $\lambda_2$ of the graph Laplacian $\L$ (the first eigenvalue $\lambda_1$ of $\L$ is always $0$).
The second smallest eigenvalue $\lambda_2$ is called the \textit{Fiedler Number}---also called the algebraic connectivity. 
The second eigen-pair $(\lambda_2, \v_2)$ of $\L$ can be computed using a known extreme eigen-pair computation algorithm such as \textit{locally optimal block preconditioned conjugate gradient} (LOBPCG) \cite{knyazev01} in linear time $\cO(N)$.
Given a complete graph $\cG$, we can remove an edge whose removal would induce the minimum change in $\lambda_2$ per iteration. 
We call this the \textit{Fiedler method}.

\vspace{-0.05in}
\subsection{Comparison of Graph Sparsification Methods}

\vspace{-0.05in}
We compare the performance of different graph sparsification schemes described in Section\;\ref{subsec:sparseMethod} in GCN training cost function value. 
Experimental details are discussed in Section\;\ref{sec:Results}. 
Due to the high complexity of APSP, we constructed the graphs only for $11$ counties in the state of Iowa. 
We inputted our computed graphs $\cG^*$ together with $11$ features (from the target year) related to crop yields. 
%such as \textit{normalized difference vegetation index} (NDVI) and \textit{enhanced vegetation index} (EVI)~\cite{matsushita2007} for GCN training. 

\begin{figure}[t]
\centering
\hspace{-0.1in}
\includegraphics[width=1\columnwidth]{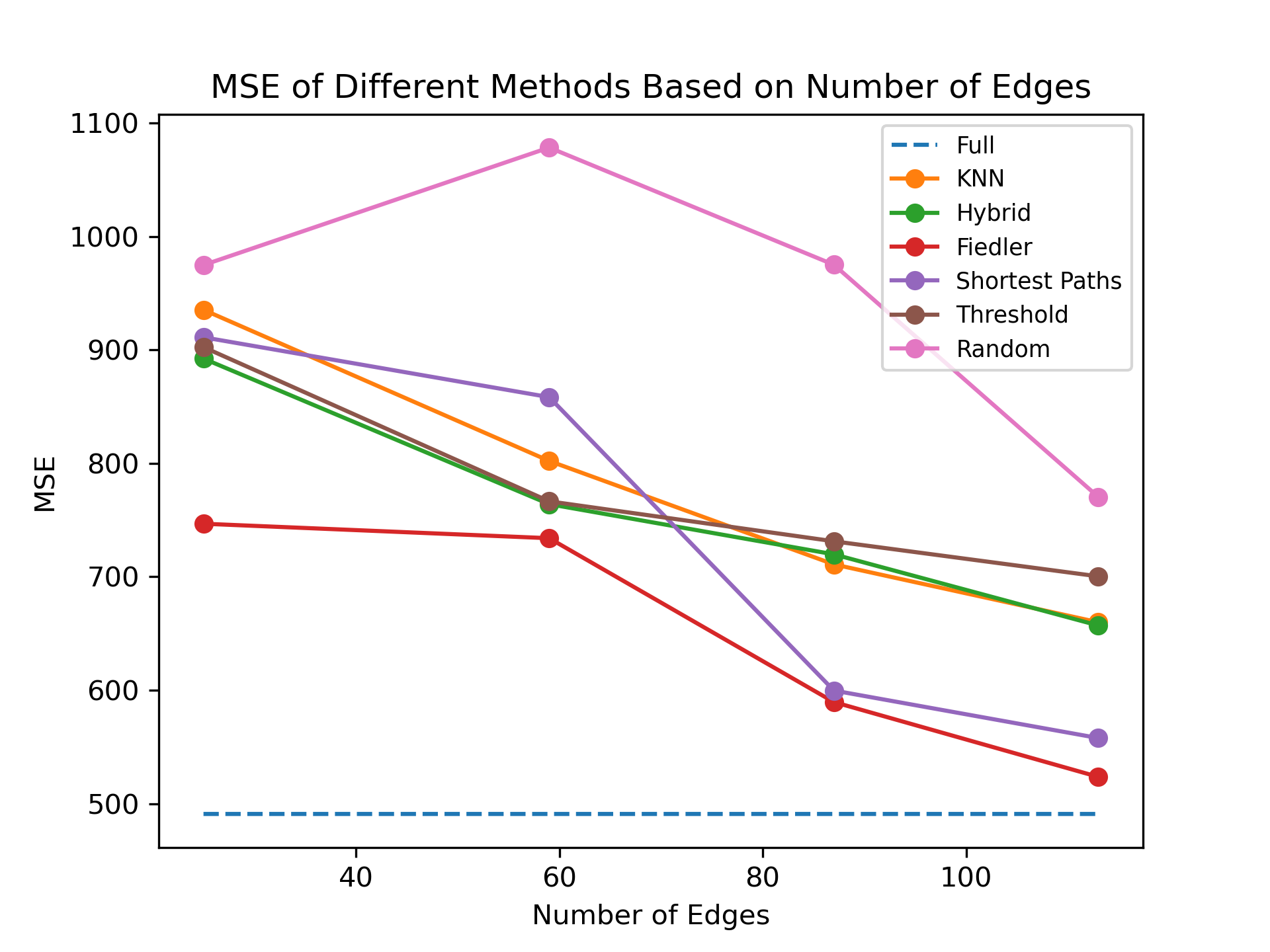}
\vspace{-0.2in}
\caption{Crop yield prediction average error in 20 runs (MSE) for competing sparsification methods compared to the full graph with 11 nodes.}
\label{fig:result_11_mse}
\end{figure}

Fig.\;\ref{fig:result_11_mse} shows the result of the six methods for four different edge counts in terms of mean square error (MSE).
Results demonstrate that the Fiedler sparsification method has the best prediction result for all sparsity counts.

\vspace{-0.05in}
\subsection{Fast Graph Sparsification}

\vspace{-0.05in}
Though the Fiedler method yields high performance, the computation complexity to select edges for removal is high. 
For each edge removal from a graph $\cG$ with $M$ edges, second eigenvalue $\lambda_2$ must be computed (using LOBPCG) for the resulting Laplacian $\L$ of each one of $M$ candidates---with complexity $\cO(N)$---resulting in complexity $\cO(M N)$. 
To remove $K$ edges, the complexity is $\cO(K M N)$.

We propose a fast selection scheme approximating the Fiedler method.
Consider a modified Laplacian matrix $\tilde{\L} = \L + \E^{m,n}$ with an edge $(m,n)$ removed from original graph $\cG$ described by Laplacian $\L$, where \textit{perturbation matrix} $\E^{m,n} \in \mathbb{R}^{N \times N}$ is defined as

\vspace{-0.1in}
\begin{small}
\begin{align}
E^{m,n}_{i,j} &= \left\{ \begin{array}{ll}
w_{m,n} & \mbox{if}~~ i=m, j=n ~~\mbox{or}~~ i=n, j=m \\
-w_{m,n} & \mbox{if}~~ i=j=m ~~\mbox{or}~~ i=j=n \\
0 & \mbox{o.w.}
\end{array} \right. .
\label{eq:matrixE}
\end{align}
\end{small}

Denote by $(\lambda_2,\v_2)$ the second eigen-pair for matrix $\L$, \ie, $\L \v_2 = \lambda_2 \v_2$.
A known eigenvalue perturbation theorem \cite{ipsen09} states that there exists an eigenvalue $\tilde{\lambda}$ of $\tilde{\L}$ in the range:
\begin{align}
\tilde{\lambda} \in \left[~ \lambda_2 - \|\E^{m,n} \v_2\|_2, ~\lambda_2 + \|\E^{m,n} \v_2\|_2 ~ \right] .
\label{eq:eVal_bound}
\end{align}

Recall in Section\;\ref{subsubsec:Fiedler} that the Fiedler method removes an edge whose removal induces the minimum change in $\lambda_2$. 
Thus, from \eqref{eq:eVal_bound}, we can derive a criterion to choose an edge $(m,n)$ for removal so that $\tilde{\lambda}$ differs from $\lambda_2$ minimally.
Denote by $\cE$ the edge set in graph $\cG$ specified by Laplacian $\L$.
We choose an edge $(m^*,n^*) \in \cE$ such that
\begin{align}
(m^*,n^*) = \arg \min_{(m,n)\in\cE} ~ \left\| \E^{m,n} \v_2 \right\|_2 .
\label{eq:approx}
\end{align}
The complexity of computing $\|\E^{m,n} \v_2\|_2$ for a given $\E^{m,n}$ is $\cO(1)$; by \eqref{eq:matrixE}, there are only four non-zero entries in $\E^{m,n}$. 
Thus, the complexity of computing $(m^*,n^*)$ in \eqref{eq:approx} is $\cO(M)$. 
If $K$ edges are removed, then the complexity is $\cO(K M)$. 

\begin{figure}[t]
\centering
\hspace{-0.1in}
\includegraphics[width=1\columnwidth]{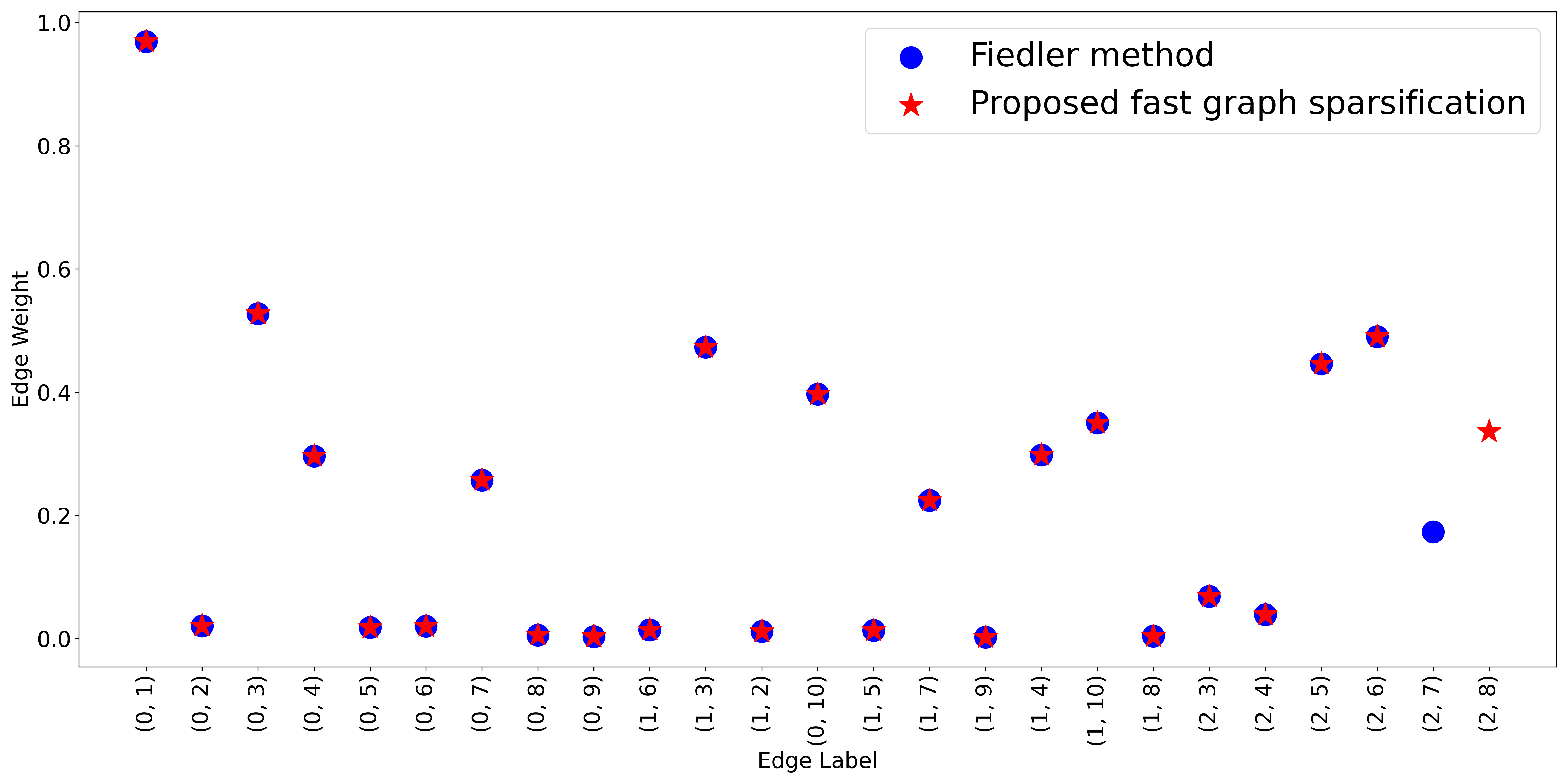}
\vspace{-0.2in}
\caption{Comparison of removed edges via the Fiedler method compared to our proposed fast approximation.}
\label{fig:result_edge_compare}
\end{figure}

Fig.\;\ref{fig:result_edge_compare} shows the removed edges (48 edges removed from 121 edges of the full graph with 11 nodes) by the Fiedler method in comparison with our proposed fast approximation. 
We observe that these two methods choose almost the same set of edges for removal. We observe also that these methods mostly remove edges with smaller weights.  

\vspace{-0.05in}
\section{Experimental Results}
\label{sec:Results}
\vspace{-0.05in}
To test the effectiveness of our proposed graph sparsification method, we conducted the following experiment. 
We used the county-level corn yield data from 2010 to 2020 provided by USDA and the National Agricultural Statistics Service\footnote{https://quickstats.nass.usda.gov/} to predict yields in 2021. 
We conducted our experiments in $99$ counties in the state of Iowa.
We first constructed a full graph based on the average yield of the previous four years and the locations of the counties. Specifically, weight $w_{i,j}$ of edge $(i,j)$ connecting counties $i$ and $j$ is computed as
\begin{align}
w_{i,j} = \exp \left( -\frac{\|\l_i - \l_j\|^2_2}{\sigma_l^2} - \frac{|y_i - y_j|^2}{\sigma_y^2} \right)
\label{eq:edgeWeight}
\end{align}
where $\l_i \in \mathbb{R}^2$ and $\y_i$ are the 2D coordinate and \textit{crop yield mean} of county $i$, respectively. %and $\sigma_l$ and $\sigma_y$ are parameters. 
We chose parameters $\sigma_l = 5.625 \times 10^{9}$ and $\sigma_y = 0.5$ for optimal performance.
We then used five different methods to sparsify the full graph: kNN, $\epsilon$-thresholding, hybrid kNN / $\epsilon$-thresholding, random, and our proposed fast Fiedler method. 
We tested five different sparsity counts (number of graph edges) for all methods. 
%These five sparsity counts were selected based on different values for $k$ in the kNN method: $k \in \{10, 25, 40, 65, 80\}$.

We inputted our constructed graphs $\cG^*$ together with $41$ features (from the target year) related to crop yields such as precipitation, soil composition and sunlight condition to the network. 
% Figure\;\ref{fig:algo_illustration}\red{add plot} shows the schematic illustration of our method. 
We trained a GCN
model EvolveGCN-H \cite{pareja19} for yield prediction. 
EvolveGCN-H combines the simple GCN \cite{kipf17} with a \text{gated recurrent unit (GRU)} \cite{cho14}. We only used one layer of EvolveGCN-H and one linear layer to avoid over-fitting. 
We calculated the average error over 20 runs for all the methods. In each run, we used 1000 epochs in addition to early stopping with patience $20$. 
We chose a learning rate $0.01$ for all our training.

\begin{figure}[t]
\centering
\includegraphics[width=1\columnwidth]{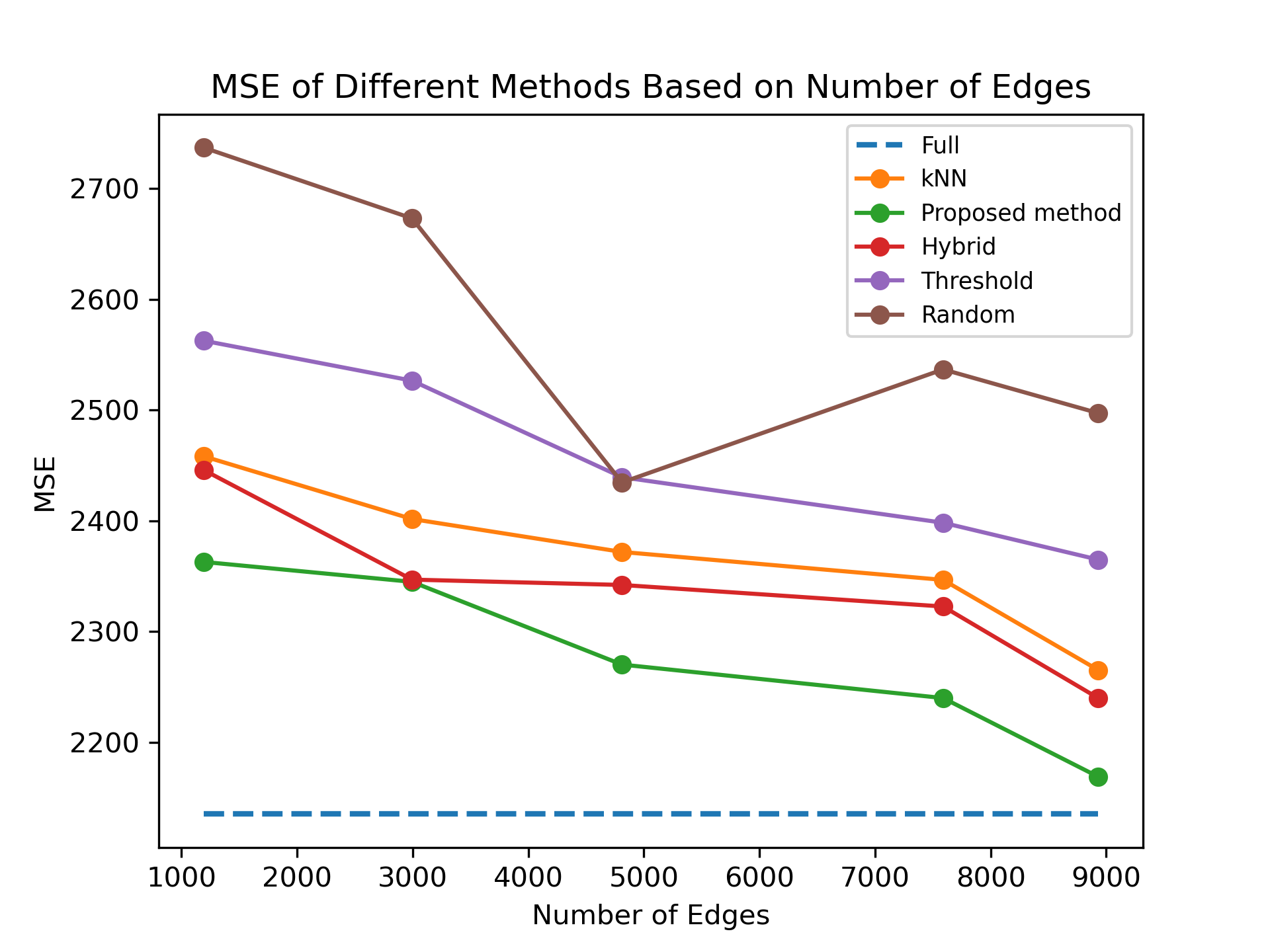}
\vspace{-0.35in}
\caption{Crop yield prediction average error (MSE) in 20 runs for five competing graph sparsification methods compared to the full graph with 99 nodes.}
\label{fig:result_Iowa_counties}
\end{figure} 

Fig.\;\ref{fig:result_Iowa_counties} shows the result of the five methods in terms of MSE.
Results demonstrate that our fast Fiedler method has the best prediction result for almost all sparsity counts. 
We observe that MSE for all methods decreased when the number of edges increased, except for random. 
With more edges, sparse graphs had more informative edge weights, hence they performed closer to the full graph.

Finally, to highlight the speed advantage of sparse graphs over full graphs during GCN training,  we recorded the average time and number of epochs in Table\;\ref{tab:table_time}. 
For this experiment, sparse graphs contained 1191 edges. 
The results demonstrate that graph sparsification led to GCN convergence in fewer number of epochs, reducing training time. 
%This benefit is particularly notable when dealing with GCN training that involves a large number of nodes.

\begin{table}[h!] \centering
\caption{Average time and number of epochs in 20 runs for four competing graph sparsification methods compared to the full graph with 99 nodes.}
\label{tab:table_time}
 \resizebox{0.88\textwidth}{!}{\begin{minipage}{\textwidth}
\begin{tabular}{|l|l|l|l|l|l|}
    \hline
    \multirow{1}{*}{Graph} &
    Full &
     kNN &
      Hybrid &
      Fast Fiedler &
     $\epsilon$-Threshold \\
    \hline
    Time & 52.70s & 17.33s &  18.08s & 15.01s & 15.70s \\
    \hline
    Epochs & 765 & 256 & 383 & 258 & 304\\
    \hline
\end{tabular}
  \end{minipage}}
\end{table}

\vspace{-0.1in}
\section{Conclusion}
\label{sec:Conclusion}
\vspace{-0.05in}
While graph convolutional nets (GCN) are effective in predicting future crop yields given relevant features and crop yields of previous years as input, an informative but dense graph kernel leads to large training and execution costs.
In this paper, we propose a fast graph sparsification method based on the Fiedler number, where an edge's removal inducing the minimal change in the second eigenvalue is chosen per iteration.
A fast version of the method leverages an eigenvalue perturbation theorem, so that the complexity of edge selection per iteration is only $\cO(M)$ for $M$ edges.
Experimental results show that our Fiedler-based sparsification method can significantly reduce GCN training time while retaining good crop yield prediction performance.

%\newpage
\begin{small}
\bibliographystyle{IEEEbib}
\bibliography{refs_full}

\begin{thebibliography}{10}

\bibitem{Cai17}
Yiqing Cai, Ayman Elhaddad, Jerrod Lessel, Christianna Townsend, Yi~Zhao, and
  Nemo Semret,
\newblock ``{Crop Yield Predictions - High Resolution Statistical Model for
  Intra-season Forecasts Applied to Corn in the US},''
\newblock in {\em AGU Fall Meeting Abstracts}, Dec. 2017, vol. 2017, pp.
  GC31G--07.

\bibitem{Sun19}
Jie Sun, Liping Di, Ziheng Sun, Yonglin Shen, and Zulong Lai,
\newblock ``County-level soybean yield prediction using deep cnn-lstm model,''
\newblock {\em Sensors}, vol. 19, no. 20, 2019.

\bibitem{Shahhosseini21}
Mohsen Shahhosseini, Guiping Hu, Saeed Khaki, and Sotirios~V. Archontoulis,
\newblock ``Corn yield prediction with ensemble {CNN}-{DNN},''
\newblock {\em Frontiers in Plant Science}, vol. 12, aug 2021.

\bibitem{kipf17}
Thomas~N. Kipf and Max Welling,
\newblock ``Semi-supervised classification with graph convolutional networks,''
\newblock in {\em International Conference on Learning Representations}, 2017.

\bibitem{oono2020graph}
Kenta Oono and Taiji Suzuki,
\newblock ``Graph neural networks exponentially lose expressive power for node
  classification,''
\newblock in {\em International Conference on Learning Representations (ICLR)},
  2020.

\bibitem{zeng23}
Jin Zeng, Yang Liu, Gene Cheung, and Wei Hu,
\newblock ``Sparse graph learning with spectrum prior for deep graph
  convolutional networks,''
\newblock in {\em IEEE ICASSP}, 2023, pp. 1--5.

\bibitem{ortega18ieee}
Antonio Ortega, Pascal Frossard, Jelena Kovacevic, Jose M.~F. Moura, and Pierre
  Vandergheynst,
\newblock ``Graph signal processing: Overview, challenges, and applications,''
\newblock {\em Proceedings of the IEEE}, vol. 106, no. 5, pp. 808--828, 2018.

\bibitem{cheung18}
Gene Cheung, Enrico Magli, Yuichi Tanaka, and Michael~K. Ng,
\newblock ``Graph spectral image processing,''
\newblock {\em Proceedings of the IEEE}, vol. 106, no. 5, pp. 907--930, 2018.

\bibitem{hu20}
Wei Hu, Xiang Gao, Gene Cheung, and Zongming Guo,
\newblock ``Feature graph learning for 3d point cloud denoising,''
\newblock {\em IEEE Transactions on Signal Processing}, vol. 68, pp.
  2841--2856, 2020.

\bibitem{yang22}
Cheng Yang, Gene Cheung, and Wei Hu,
\newblock ``Signed graph metric learning via {Gershgorin} disc perfect
  alignment,''
\newblock {\em IEEE Transactions on Pattern Analysis and Machine Intelligence},
  vol. 44, no. 10, pp. 7219--7234, 2022.

\bibitem{Chung1997}
Fan R.~K. Chung,
\newblock {\em Spectral Graph Theory},
\newblock American Mathematical Society, Providence, RI, 1997.

\bibitem{ipsen09}
Ilse Ipsen and Boaz Nadler,
\newblock ``Refined perturbation bounds for eigenvalues of hermitian and
  non-hermitian matrices,''
\newblock {\em SIAM J. Matrix Analysis Applications}, vol. 31, pp. 40--53, 01
  2009.

\bibitem{floyd62}
Robert~W. Floyd,
\newblock ``Algorithm 97: Shortest path,''
\newblock {\em Commun. ACM}, vol. 5, no. 6, pp. 345, jun 1962.

\bibitem{knyazev01}
Andrew~V. Knyazev,
\newblock ``Toward the optimal preconditioned eigensolver: Locally optimal
  block preconditioned conjugate gradient method,''
\newblock {\em SIAM Journal on Scientific Computing}, vol. 23, no. 2, pp.
  517--541, 2001.

\bibitem{pareja19}
Aldo Pareja, Giacomo Domeniconi, Jie Chen, Tengfei Ma, Toyotaro Suzumura,
  Hiroki Kanezashi, Tim Kaler, Tao~B. Schardl, and Charles~E. Leiserson,
\newblock ``{EvolveGCN}: Evolving graph convolutional networks for dynamic
  graphs,''
\newblock in {\em Proceedings of the Thirty-Fourth AAAI Conference on
  Artificial Intelligence}, 2020.

\bibitem{cho14}
Kyunghyun Cho, Bart van Merri{\"e}nboer, Dzmitry Bahdanau, and Yoshua Bengio,
\newblock ``On the properties of neural machine translation: Encoder{--}decoder
  approaches,''
\newblock in {\em Proceedings of {SSST}-8, Eighth Workshop on Syntax, Semantics
  and Structure in Statistical Translation}, Doha, Qatar, Oct. 2014, pp.
  103--111, Association for Computational Linguistics.

\end{thebibliography}
\end{small}

\end{document}